# Introducing δ -XAI: a novel sensitivity-based method for local AI explanations

Alessandro De Carlo[*a], Enea Parimbelli[a,b], Nicola Melillo[+a], Giovanna Nicora[+a]

[a] Department of Electrical, Computer and Biomedical Engineering, University of Pavia

[B] Telfer school of management, University of Ottawa

[*]corresponding author, [+]contributed equally

# Abstract

Explainable Artificial Intelligence (XAI) is central to the debate on integrating Artificial Intelligence (AI) and Machine Learning (ML) algorithms into clinical practice. High-performing AI/ML models, such as ensemble learners and deep neural networks, often lack interpretability, hampering clinicians' trust in their predictions. To address this, XAI techniques are being developed to describe AI/ML predictions in human-understandable terms. One promising direction is the adaptation of sensitivity analysis (SA) and global sensitivity analysis (GSA), which inherently rank model inputs by their impact on predictions. Here, we introduce a novel δ-XAI method that provides local explanations of ML model predictions by extending the δ index, a GSA metric. The δ-XAI index assesses the impact of each feature's value on the predicted output for individual instances in both regression and classification problems. We formalize the δ-XAI index and provide code for its implementation. The δ-XAI method was evaluated on simulated scenarios using linear regression models, with Shapley values serving as a benchmark. Results showed that the δ-XAI index is generally consistent with Shapley values, with notable discrepancies in models with highly impactful or extreme feature values. The δ-XAI index demonstrated higher sensitivity in detecting dominant features and handling extreme feature values. Qualitatively, the δ-XAI provides intuitive explanations by leveraging probability density functions, making feature rankings clearer and more explainable for practitioners. Overall, the δ-XAI method appears promising for robustly obtaining local explanations of ML model predictions. Further investigations in real-world clinical settings will be conducted to evaluate its impact on AI-assisted clinical workflows.

# Introduction

Artificial Intelligence (AI) and Machine Learning (ML) have advanced significantly, from rule-based systems to deep learning techniques, ultimately leading to Foundation Models' development [1]. However, the application of these methods in high-stakes fields like medicine remains confined to research settings [2,3]. This limitation is partly due to the potential lack of transparency of some AI methods, such as deep learning, whose internal reasoning process is often obscure. Many of the most performing AI algorithms, including deep networks and ensembles, are considered "black boxes", as they prevent human users from understanding the model's behavior during classification [4,5]. Various methods in the realm of "Explainable AI" (XAI) have been developed to increase the interpretability of black box models. These approaches aim to provide a *global* or *local* understanding of the classifier's reasoning process. Global explainability focuses on the overall impact of the features on the model's predictions, while local explainability refers to the ability to explain individual predictions *case by case*. Local XAI offers explanations that can enhance understanding of feature contributions within smaller groups of individuals often overlooked by global interpretation techniques [6]. Many XAI methods are now available. Approaches like LIME provide local explanations by approximating complex models with simpler, linear models (explainable by design) on a neighborhood of the examples whose prediction needs to be explained [7,8]. The strong assumption of these methods is that a simple linear model can be a proper proxy for a complex classifier *locally*. Conversely, SHAP is a local XAI method based on the game's optimal Shapley values. SHAP assigns to each feature a score (i.e. a Shapley value) that should reflect the importance of the specific feature for the prediction [9]. SHAP has been widely used and several extensions were proposed [10]. Yet, some works have identified issues in SHAP explanations when Shapley values are used for feature importance [11], and some corrections have been proposed [12].

Sensitivity Analysis (SA) and Global Sensitivity Analysis (GSA) investigate how the variation of the model inputs influences the model output predictions [13–15]. Generally, the sources of variation can be related to the uncertainty of the estimation process (e.g., model inputs are parameters identified on a data set) or to the variability of the input in a population (e.g., model inputs are features characterizing each individual) [16]. GSA is a branch of SA that relies on the multivariate variation of all the model inputs of interest [13]. GSA can be defined as *"the study of how uncertainty in the output of a model can be apportioned to different sources of uncertainty in the model input"* [13]. GSA methods allow to rank model inputs (called *'factors'* in the GSA jargon) according to their impact on the model output variation and to identify the model parameters whose uncertainty/variability should be reduced to obtain more reliable model predictions. GSA is usually applied in mechanistic and statistical modelling of human and natural systems [15], and several methods have been proposed in the literature [17,18]. The most established and widely used is the variance-based GSA, where factors are ranked according to their contribution to the variance decomposition of the model output [19,20]. Differently, moment-independent GSA techniques consider the entire distribution of the output rather than a single statistical moment (e.g., variance) [15,17,27]. In particular, the $\delta$ sensitivity index [28] describes the impact of each model parameter on the output probability density function. The $\delta$ sensitivity index was reported to provide robust results, independently from the shape of the output distribution. Conversely, variance-based methods can be misleading when the output distribution is multi-modal [27] or highly-skewed [29], since variance is a sensible measure of the output variation [30]. In addition, δ index is well-defined also in the presence of statistical dependencies between the model inputs [28], a situation widely encountered in machine learning when training on real-world data [31].

The ability to decompose the model output being robust to feature correlations and data distributions, makes the δ sensitivity index particularly suitable for XAI. Recently, GSA has been proposed to provide global explanations of ML predictions [32]. However, we believe that the δ sensitivity index can be highly useful to provide local explanations as well.

Therefore, the aim of this work is to introduce a novel local explainability method for AI models, based on the δ sensitivity index, called δ-XAI. Its [tical basis ensures it remains robust to features correlation, and across different distributions. More in detail, the derivation of δ-XAI index is initially presented. Then, δ-XAI and Shapley values performances are compared on simple models interpretable by design. Code is available at https://github.com/bmi-labmedinfo/deltaXAI.

# Methods

## The δ GSA index

Let us consider a generic model

$$Y = g(\boldsymbol{X})$$

*( 1 )*

where $Y$ is the scalar output of the model, $g$ represents the inputs-output relationship and $\boldsymbol{X} = \{X_1, \dots, X_N\}$ is the $R^N$ vector of the model inputs. Within the GSA framework, $\boldsymbol{X}$ is considered as a random variable [13] which is characterized by a joint probability density function (pdf), $f(\boldsymbol{X})$. Therefore, $Y$ is a random variable with a pdf, $f(Y)$, which can be calculated through Eq.1 using input samples extracted from $f(\boldsymbol{X})$.

The definition of the δ index relies on the following considerations [28]. Suppose that one input $X_i$ can be fixed to a certain value $x_i^*$, then, the conditional pdf of $Y$ given $X_i = x_i^*$, $f(Y|X_i = x_i^*)$, can be defined. The shift between $f(Y)$ and $f(Y|X_i = x_i^*)$ can be measured as

$$s(X_i) = \int |f(Y) - f(Y|X_i = x_i^*)| dY.$$

*(2)*

As illustrated in Figure 1, $s(X_i)$ represents the difference of the area underlying $f(Y)$ and $f(Y|X_i = x_i^*)$, which corresponds to the impact of fixing $X_i$ to $x_i^*$ on $f(Y)$. $X_i$ is a random variable typically assuming more values than just $x_i^*$. The $\delta$ sensitivity index for $X_i$ can be computed through the expected value of $s(X_i)$ over the entire domain of $X_i$ as in Eq.3, where $f(X_i)$ is the marginal pdf of $X_i$.

$$\delta_i = \frac{1}{2} E_{X_i}[s(X_i)] = \int f(X_i) \left[ \int |f(Y) - f(Y|X_i)| \, dY \right] dX_i.$$

( 3 )

It has been demonstrated in [28] that $0 \leq \delta_i \leq 1$, in particular $\delta_i = 0$ if and only if $Y$ is independent from $X_i$ and $X_i$ is uncorrelated from the other $X_j$, with $i \neq j$ A full description of the properties of this sensitivity index can be found in [28].

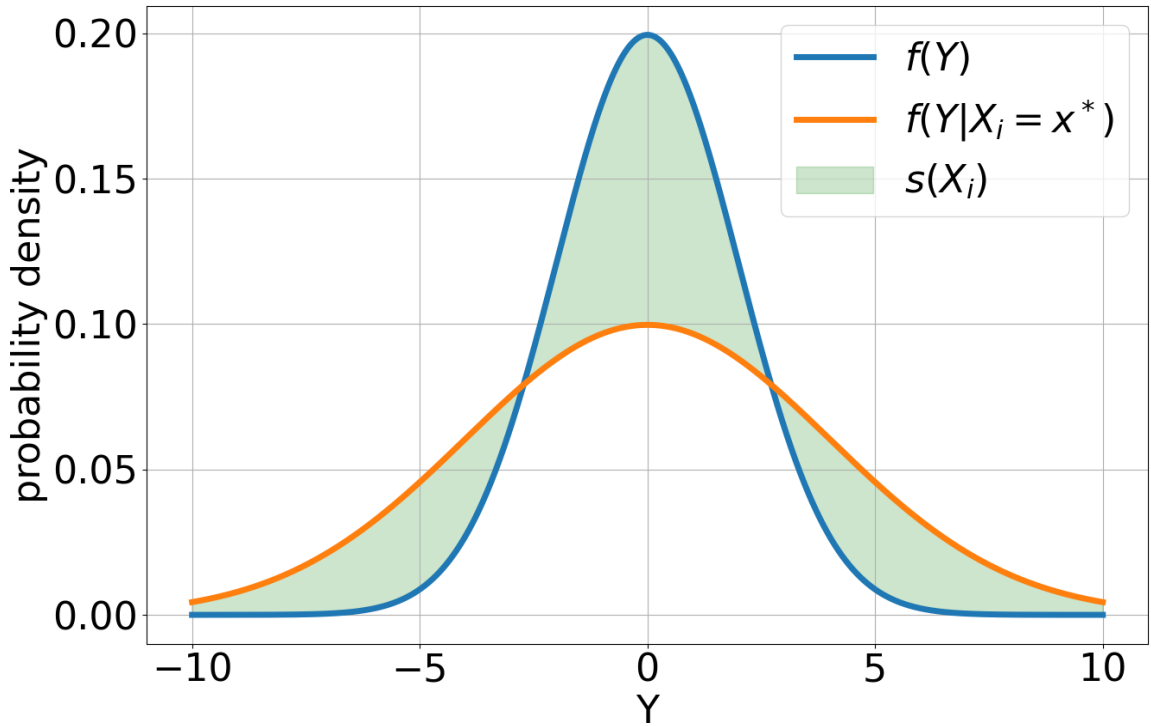

**Figure 1** Comparison between conditional and unconditional pdf of Y. Green shaded area quantifies the impact of $X_i = x_i^*$ on $f(Y)$.

# Adaptation of the δ GSA index for local explainability of ML models: the δ-XAI index

Let us introduce $h$, a generic ML model taking as input a generic $\mathbb{R}^M$ features vector characterizing each instance/example $\boldsymbol{x} = \{x_1, \dots, x_M\}$, and returning a scalar prediction, $y = h(\boldsymbol{x})$, with $y \in \mathbb{R}$. Consider also the training set of $h$ containing $N$ examples, $\boldsymbol{X} \in \mathbb{R}^{N \times M}$, representing the knowledge on the domain of interest (e.g., sample from a population of interest). The concepts of the GSA can be easily mapped within this framework. Indeed, each feature, $x_i$, has a variability in the domain of interest, therefore a joint probability density function (i.e., pdf) for the features, $f(\boldsymbol{x})$, can be defined. Consequently, the model output is also random variable with a pdf, $f(y)$, describing the variation of the predicted variable within its domain.

Given a particular example, $\boldsymbol{x}^* = \{x_1^*, \ldots, x_M^*\}$, which is assumed to be drawn from $f(\boldsymbol{x})$, it is possible to compute the model output, $y^* = h(\boldsymbol{x}^*)$, and its probability density within the domain of the predicted variable, $f(y = y^*)$. Analogously to Eq.2, the impact of each feature value, $x_i = x_i^*$, on the final model prediction, $y^* = h(\boldsymbol{x}^*)$, can be obtained from Eq.4

$$\delta_i = f(y = y^* | x_i = x_i^*) - f(y = y^*).$$

( 4 )

In particular, $f(y = y^* | x_i = x_i^*)$ is the probability density of $y^*$ conditioned by having observed $x_i = x_i^*$. Therefore, $\delta_i$ is the shift in the probability density function for $y = y^*$, when the feature $x_i = x_i^*$. Indeed, as $\delta_i \to 0$, $f(y = y^* | x_i = x_i^*)$ is close to the probability density due to the variation of the other features in $\boldsymbol{X}$ (i.e., $f(y = y^*)$). Conversely, a $|\delta_i| > 0$ indicates that $x_i = x_i^*$ increases/decreases the likelihood of obtaining $y^*$.

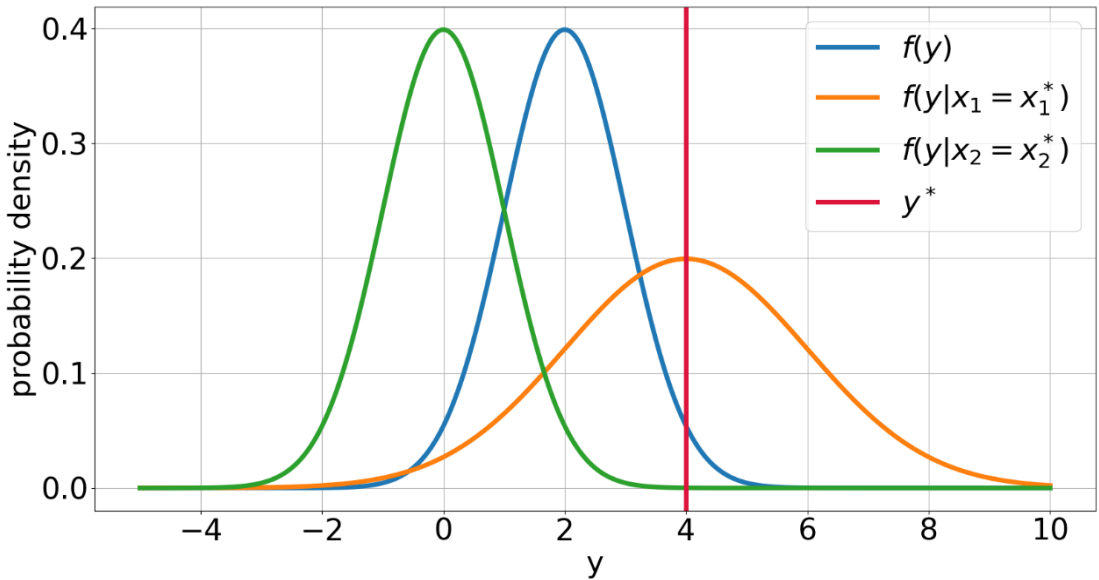

**Figure 2** A graphical example of $\delta_i$ interpretation. Blue line is the pdf of model output, y, on the entire domain. Orange and green lines are the pdf of y conditioned by the observed values for the features $x_{1=}x_1^*$ and $x_2 = x_2^*$, respectively.

Then, features can be ranked independently from the sign of $\delta_i$, by applying the normalization in Eq. 5

$$\hat{\delta}_t = \frac{|\delta_i|}{\sum_i^M |\delta_i|}.$$

( 5 )

# Extension of the δ -XAI index to binary classification problems

In the context of classification problems, given a class target, $c_t$, the ML model, $h$, returns the probability that the example $\boldsymbol{x}$ belongs to $c_t$, $y = h(\boldsymbol{x}) = p(\boldsymbol{x} \in c_t)$. Then, given a decision threshold $d_t$, $\boldsymbol{x}$ is labelled with $c_t$ whether $y \geq d_t$, otherwise with $\overline{c_t}$ (i.e., the opposite of the target class). The variation of the features in the domain of interest can be propagated on $y$ also in that case. Thus, $f(y)$, representing the pdf of $p(\boldsymbol{x} \in c_t)$ in the target domain, can be defined. However, in classification problems the focus is on the class predicted by the model rather than on the specific $p(\boldsymbol{x} \in c_t)$. Therefore, recalling the properties of pdf, given $f(y)$, the probability of assigning $c_t$ with respect to $d_t$ (i.e., the probability of obtaining a $p(\boldsymbol{x} \in c_t) > d_t, p(p(\boldsymbol{x} \in c_t) > d_t))$ a can be obtained from Eq.6

$$p(p(\boldsymbol{x} \in c_t) > d_t) = \int_{d_t}^{1} f(y) \ dy.$$

<div align="center">( 6 )</div>

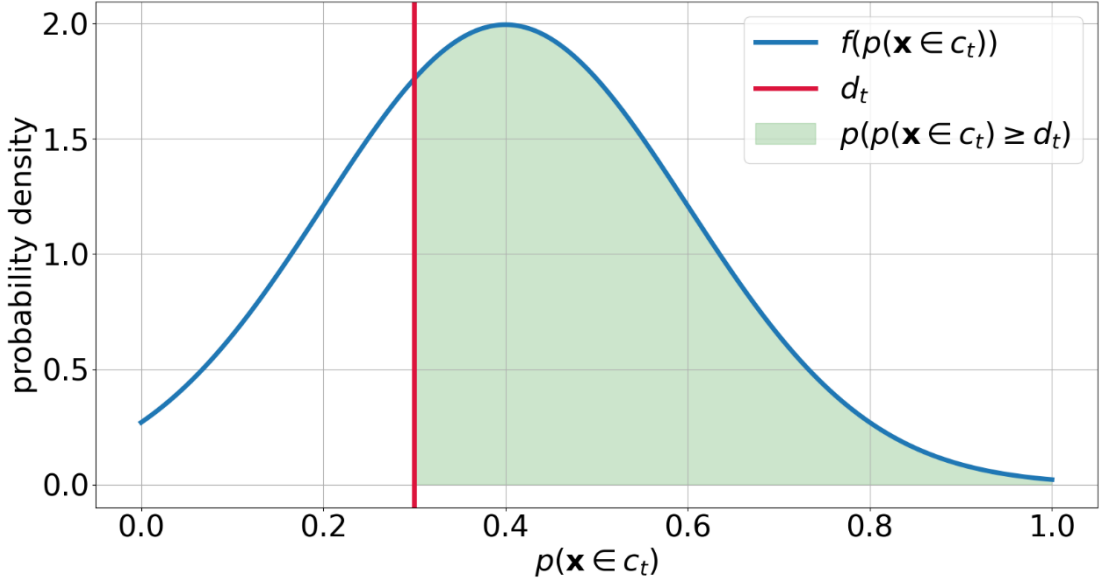

**Figure 3** Graphical representation of the scenario in the classification context. Here, the focus in on $p(p(\boldsymbol{x} \in c_t) > d_t)$ (green shaded area) rather than a specific $p(\boldsymbol{x} \in c_t)$.

In particular, Eq.6 provides the probability of assigning $c_t$ in the target domain by using $h$. Analogously, it is possible to compute this probability conditioned by the observed value for the $i - th$ feature:

$$p(p(\boldsymbol{x} \in c_t | x_i = x_i^*) > d_t) = \int_{d_t}^{1} f(y | x_i = x_i^*) \ dy.$$

<div align="center">( 7 )</div>

Consequently, to evaluate the impact of $x_i = x_i^*$ in the classification of $\boldsymbol{x}^*$, we can apply Eq. 8 representing the natural extension of Eq. to classification problems

$$\delta_i = \int_{d_t}^1 f(y|x_i = x_i^*) - f(y) \; dy.$$

( 8 )

Intuitively, $\delta_i$ quantifies how much $x_i = x_i^*$ increases/decreases the probability of assigning $c_t$ with respect to the knowledge of the target domain. More in details, a $\delta_i > 0$ indicates that $x_i = x_i^*$ pushes $h$ towards assigning $c_t$ to $x^*$. Conversely, a negative value implies that $h$ is more prone to label $x^*$ with $\overline{c}_t$. Finally, a $\delta_i$ close to 0 implies that $x_i = x_i^*$ does not impact the classification. A feature rank can be obtained by applying Eq. 5 independently from the sign of $\delta_i$ also in that case.

## Numerical implementation of the δ-XAI method

The aim of this section is to provide a description of the implemented numerical algorithm to compute the $\delta$ indices. The core of the numerical procedure is to robustly estimate pdf of the ML model output, both unconditioned and conditioned, with a data driven approach. This implicitly leads to consider training set, $X$, as a representation of knowledge characterizing the domain of interest. However, $X$ is only a sample drawn from the joint pdf of the features, $f(x)$. Therefore, it is necessary to consider sampling uncertainty when estimating $f(y)$, $f(y|x_i = x_i^*)$ and, consequently, $\delta_i$. To this end, Monte Carlo simulations based on bootstrap samplings from $X$ are leveraged to compute the $\delta$ indices (Listing 1). This approach allows to compute some statistics for each $\widehat{\delta}_i$ (i.e., median an interquartile range) describing the typical value and the variability of the ranking index. The sign of $\delta_i$ representing whether a feature value increase/decrease the probability of observing a certain model output, is established considering the signs of $\delta_i$ observed in all the Monte Carlo simulations. If more than the 95% of bootstrap samples of $\delta_i$ were positive/negative, it is possible to assign a sign to $\delta_i$. Conversely, there is not enough evidence to determine it.

**Listing 1:** Pseudo-code for the computation of the $\delta$ indices for local explainability of ML model predictions.

---

**Input:** ML model $h$, training set $X$, $L$ number of bootstrap iterations, $x^*$ instance to be predicted, $y^*$ model prediction for the targeted instance. For classification problems: class target $c_t$, decision threshold $d_t$

**loop** for each bootstrap iteration ($L$ times):

    $\widehat{X} \leftarrow$ bootstrap N examples from $X$

    $\widehat{Y} \leftarrow h(\widehat{X})$

    $K_Y \leftarrow$ kernel density estimator of model output pdf fitted on $\widehat{Y}$

    **loop** for each feature in $x^*$:

        $\widehat{X}tmp_i \leftarrow \widehat{X}$

        $\widehat{X}tmp_i[:, i] \leftarrow x_i^*$

$$\widehat{Y_{x_i}} \leftarrow h(\widehat{Xtmp}_i)$$

$K_{Y|x_i} \leftarrow$ kernel density estimator of model output pdf fitted on $\widehat{Y_{x_i}}$

Compute $\delta_i$ with Eq.4 or Eq.8

Compute $\widehat{\delta_t}$ with Eq.5

# Results

The introduced local explainability framework based on $\delta$ index was benchmarked against the current state-of-art Shapley values. Simple linear regression models were considered in this evaluation framework due to their intrinsic features ranking characteristics. Indeed, regression coefficients, $\beta_i$, provides a measure of the impact of each feature on the final prediction. In this section, the results of the performed comparisons are reported.

## Case 1

Let us consider the linear regression model in Eq.9, with all $X_i \sim N(0,1)$:

$$Y = 100 \cdot X_1 + 50 \cdot X_2 + 50 \cdot X_3.$$

(9)

Suppose that the goal is to quantify the contributes of each feature in predicting $\boldsymbol{x}^* = \{0,0,2\}$ which produces a model output $y^* = 100$. Panels A-C of Figure 4 illustrates the features ranking based on the $\delta$ index. In particular, the most impacting variable, $X_3$, has a positive contribute as it increases the probability density of observing $y^*$. $X_1$ and $X_2$ follows in the ranking, with $X_1$ being more important than $X_2$ due to its higher $\beta$ coefficient. Interestingly, $X_1 = 0$ gives a negative contribute as it lowers the probability density of having a model prediction equal to 100. Conversely, the contribute of $X_2$ is almost 0 as when it is fixed to 0 it does not significantly alter the probability density of the output (Panel C). As shown in Panel B of Figure 4, the $\delta$-based ranking is coherent with the one obtained with the Shapley values.

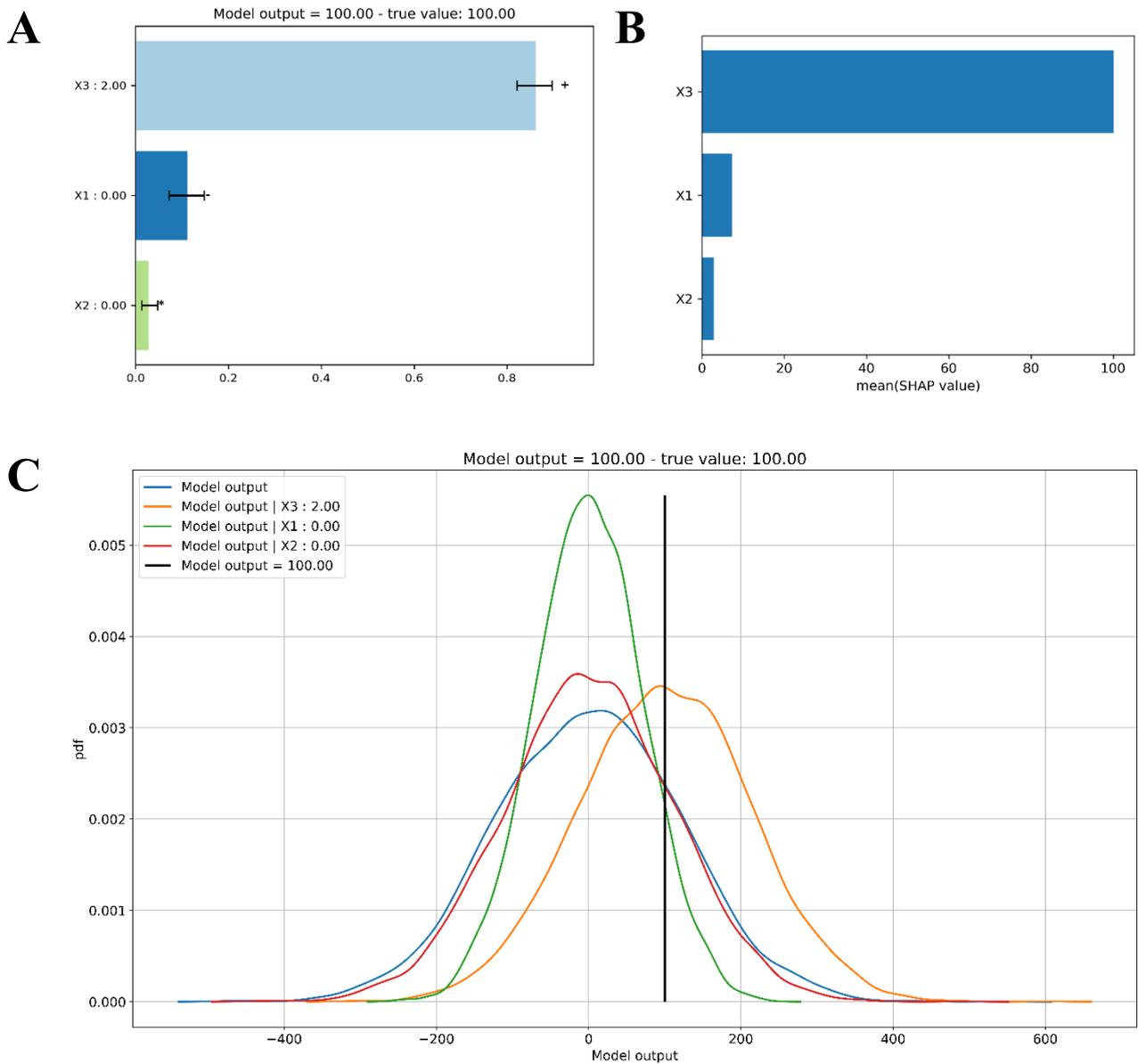

**Figure 4** Comparison between the δ indices and Shapley values on assessing features impact in predicting $\mathbf{x}^* = \{0,0,2\}$ with Eq.9. Panel A shows median and IQR (black bars) of the $\widehat{\delta_i}$ computed for each feature. A +/-/* symbol follows the bars depending by the signs of the $\delta_i$ in the computed with the bootstrap procedure. Panel B illustrates the ranking obtained with Shapley values. Panel C contains a graphical representation of how the $\delta_i$ computed for each feature value of $\mathbf{x}^*$.

Let us now consider an instance $\boldsymbol{x}^* = \{0.1, 0.1, 0.1\}$, having all features set to the same value. In this scenario, as intuitive, a feature ranking reflecting the values of the β coefficients was obtained (Figure 5).

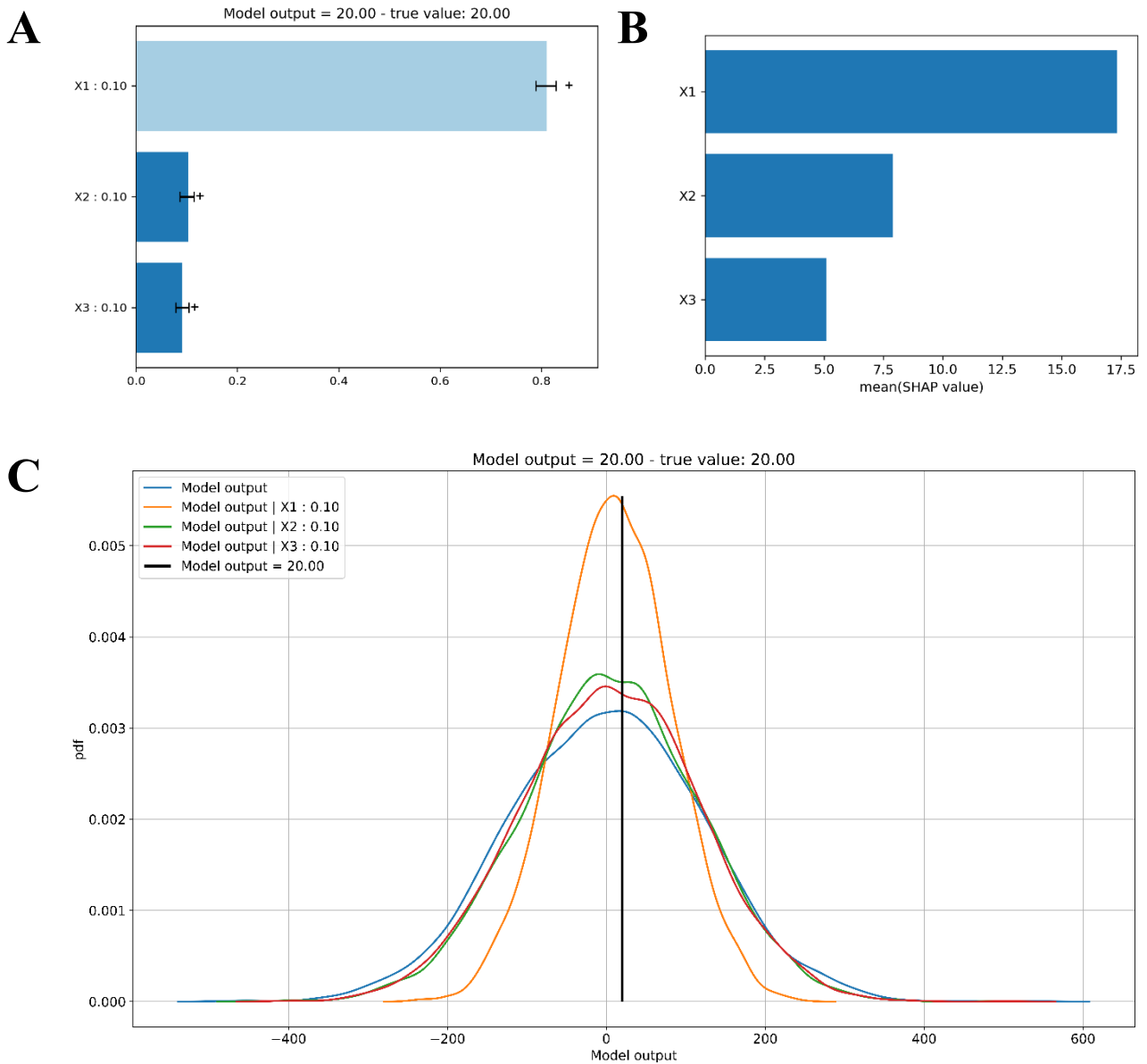

**Figure 5** Comparison between the δ indices and Shapley values on assessing features impact in predicting $\mathbf{x}^* = \{0.1, 0.1, 0.1\}$ with Eq.9. Panel A shows median and IQR (black bars) of the $\widehat{\delta_i}$ computed for each feature. A +/-/* symbol follows the bars depending by the signs of the $\delta_i$ in the computed with the bootstrap procedure. Panel B illustrates the ranking obtained with Shapley values. Panel C contains a graphical representation of how the $\delta_i$ computed for each feature value of $\mathbf{x}^*$.

When $\boldsymbol{x}^* = \{1, 2, 2\}$, the prediction of the model in Eq. 9 is 300, with each feature contributing with a value of 100 (Figure 6). The results of the Shapley values suggest a similar importance to all three features, which it seems consistent with the observations reported above (Panel B, Figure 6). Interestingly, the δ index-ranking returns $x_2 = 2$ and $x_3 = 2$ as most important features on model prediction, while $x_1 = 1$ has the lowest importance (Panel A, Figure 6). In this example, all the features follow a normal distribution with mean equal to 0 and standard deviation equal to 1. $x_2$ and $x_3$ were considered equal to 2, hence they assumed "unlikely" or "extreme" values (above the 97.5 percentile). Conversely, $x_1$ was considered equal to the more "likely" value of 1. By looking at Figure 6, panel C, it is possible to see that the output value $y = 300$ is as well a relatively "unlikely" value, being at the right tail of the output distribution. From the conditional distributions, it can be observed that fixing $x_1$ to 1, weakly impacts the probability of observing values of $y \cong 300$, with respect to the overall population. Conversely, $x_2$ or $x_3$ equal to 2, strongly impact the likelihood of observing $y \cong 300$, with respect to the overall population. The $\delta$-XAI method captures this characteristic of the model behaviour and distribution of the features, whereas the Shapley values appear to not consider that.

When considering less 'extreme' values of the features, for example $\boldsymbol{x}^* = \{0.4, 0.8, 0.8\}$, the $\delta$-XAI results are similar to the Shapley ones (Figure 7).

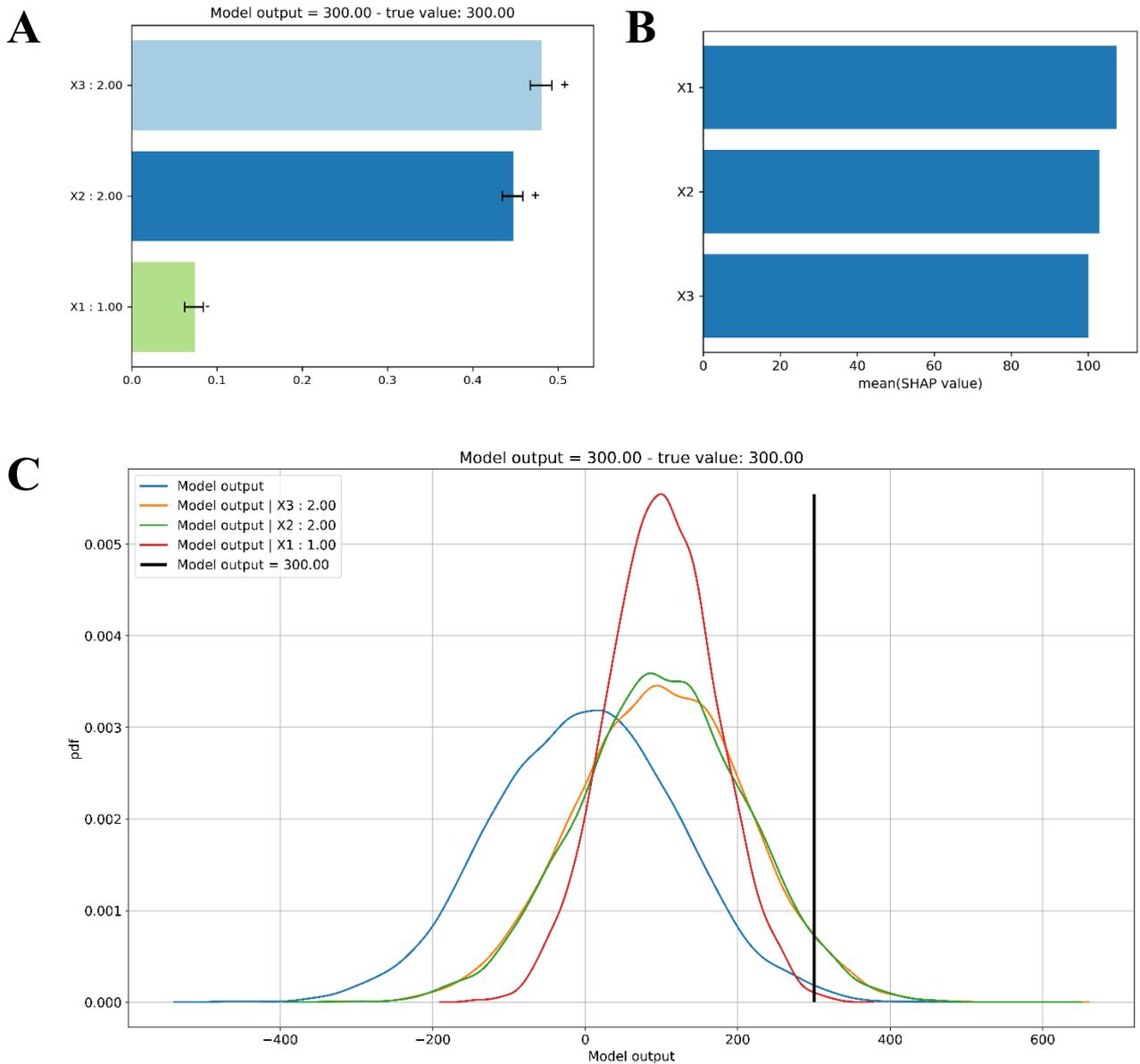

**Figure 6** Comparison between the δ indices and Shapley values on assessing features impact in predicting $\mathbf{x}^* = \{1,2,2\}$ with Eq.9. Panel A shows median and IQR (black bars) of the $\widehat{\delta_i}$ computed for each feature. A +/-/* symbol follows the bars depending by the signs of the $\delta_i$ in the computed with the bootstrap procedure. Panel B illustrates the ranking obtained with Shapley values. Panel C contains a graphical representation of how the $\delta_i$ computed for each feature value of $\mathbf{x}^*$.

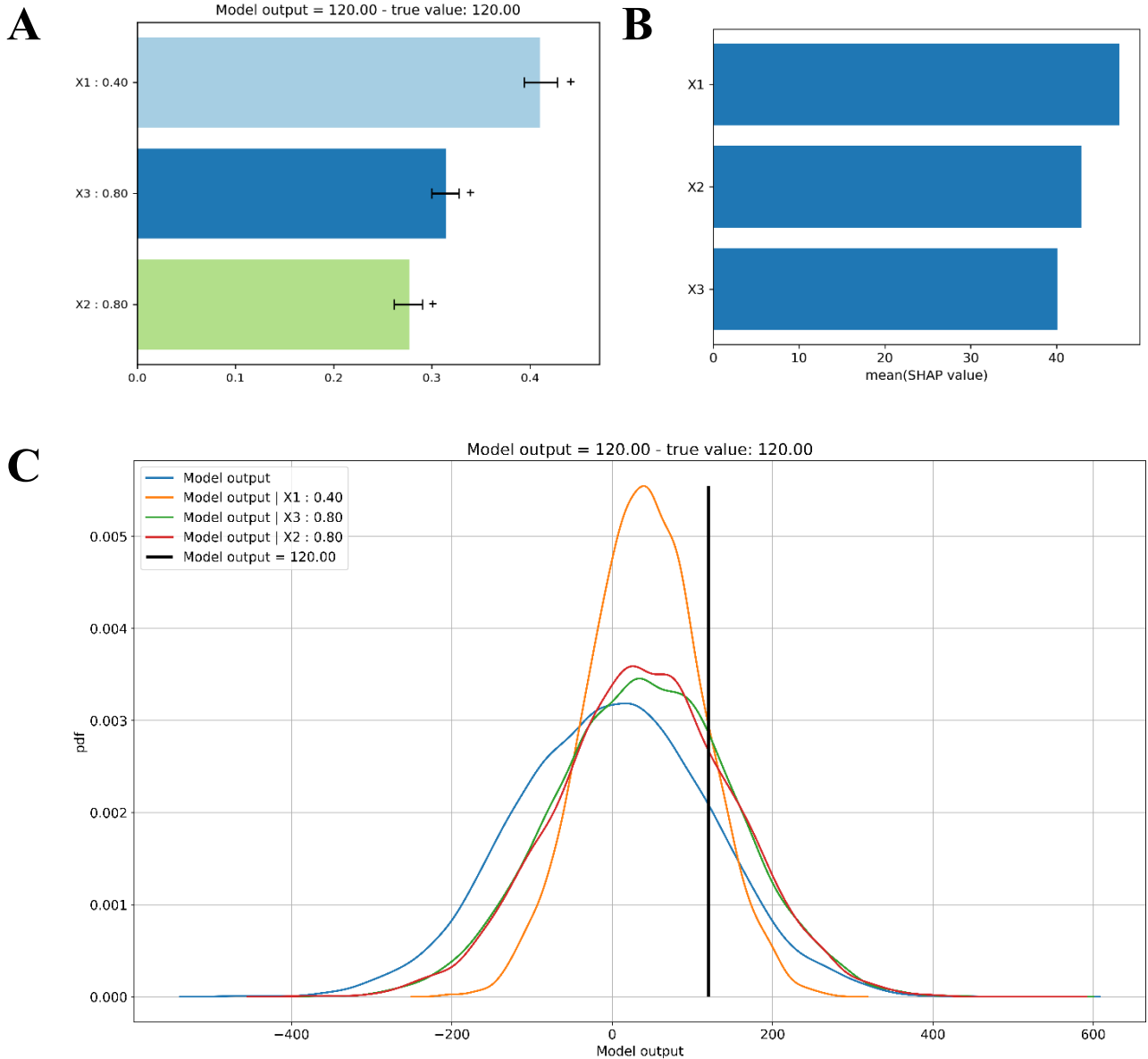

**Figure 7** Comparison between the δ indices and Shapley values on assessing features impact in predicting $\mathbf{x}^* = \{0.4, 0.8, 0.8\}$ with Eq.9. Panel A shows median and IQR (black bars) of the $\widehat{\delta_i}$ computed for each feature. A +/-/* symbol follows the bars depending by the signs of the $\delta_i$ in the computed with the bootstrap procedure. Panel B illustrates the ranking obtained with Shapley values. Panel C contains a graphical representation of how the $\delta_i$ computed for each feature value of $\mathbf{x}^*$.

## Model 2

Let us consider the linear model in Eq.10, with $X_i \sim N(0,1)$ and $X_1$ being significantly more impactful than the others due to its bigger $\beta$ coefficient.

$$Y = 1000 \cdot X_1 + 50 \cdot X_2 + 50 \cdot X_3.$$

Consider the instance $x^* = \{0,0,2\}$, already used for the model in Eq.9 (Figure 4). When Eq.10 is used to predict $x^*$, the δ index-ranking returns $x_1 = 0$ as the most impacting feature due to its dramatic impact on the probability of observing an output equal to 100 (Panel A and C, Figure 8). Differently, Shapley values give higher importance to $x_3 = 2$, i.e., the only feature different from zero (Panel B of Figure 8). It was observed that at least a $\beta = 100000$ for $X_1$ in Eq.10 brings Shapley to return a features ranking similar to those of the δ-XAI (Panel D of Figure 8).

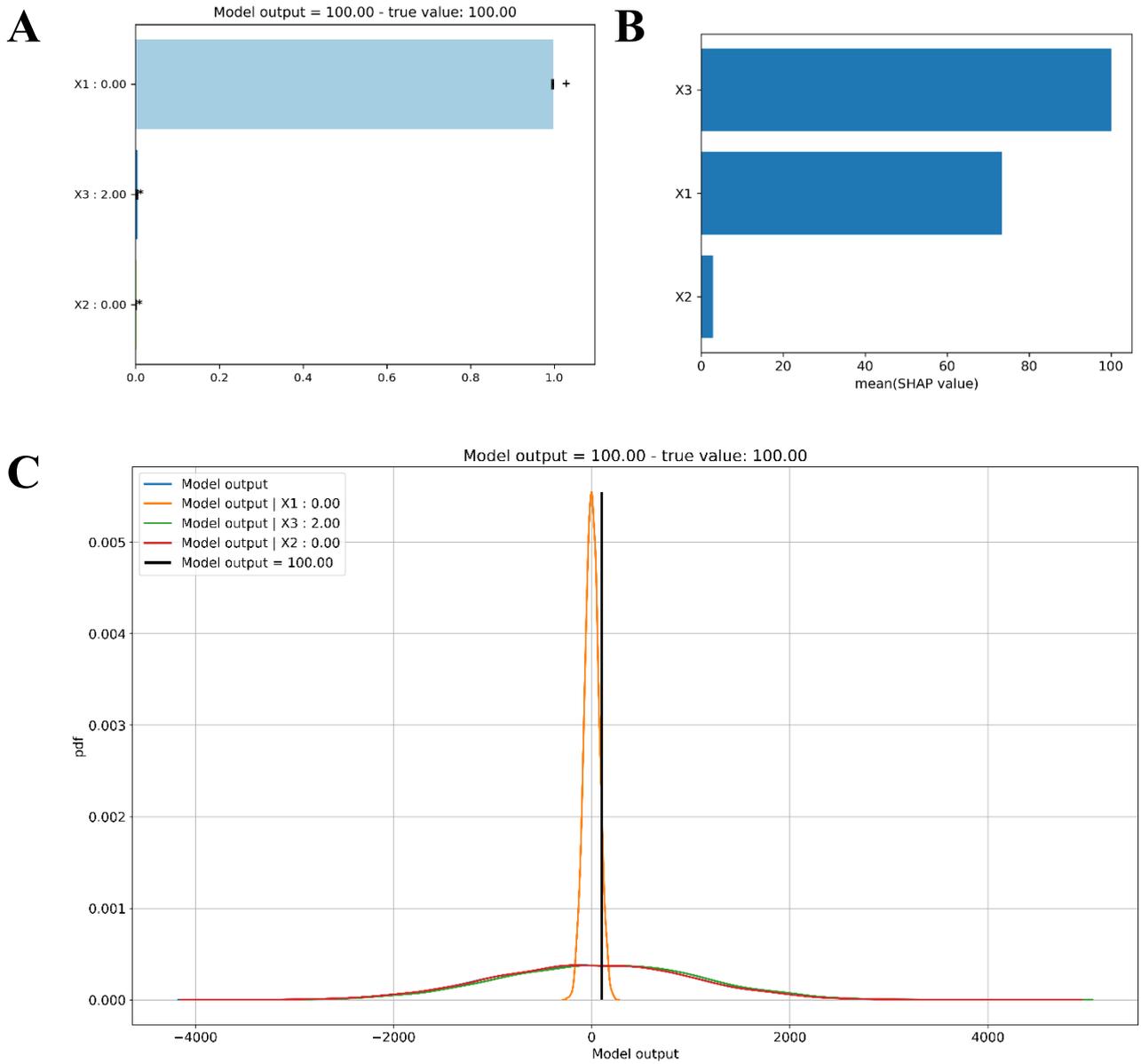

**D**

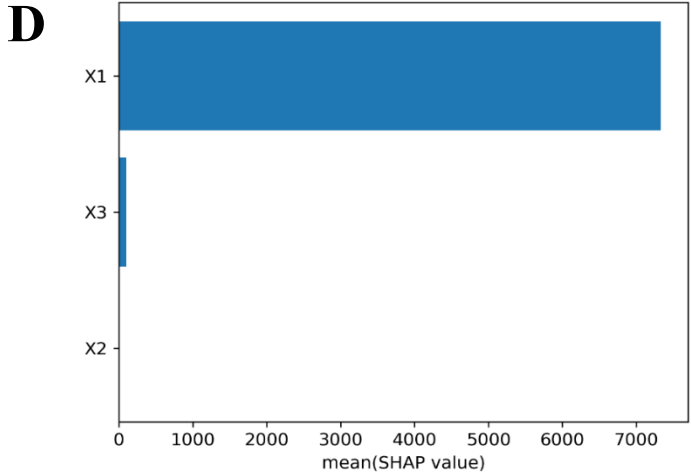

**Figure 8 7** Comparison between the δ indices and Shapley values on assessing features impact in predicting $\mathbf{x}^* = \{0,0,2\}$ with Eq.10. Panel A shows median and IQR (black bars) of the $\widehat{\delta_i}$ computed for each feature. A +/-/* symbol follows the bars depending by the signs of the $\delta_i$ in the computed with the bootstrap procedure. Panel B illustrates the ranking obtained with Shapley values. Panel C contains a graphical representation of how the $\delta_i$ computed for each feature value of $\mathbf{x}^*$. Finally, panel D shows that Shapley values can return a similar ranking to the **δ**-XAI method when the **β** of $\mathbf{X_1}$ is set to 100000.

## Impact of correlations between model features

The following examples deal with assessing the performance of δ index in the presence of correlated features. In particular, the first experiment evaluates the robustness of the methodology in when a feature not included in the model, is strongly correlated with a feature which is included in the model. To this end, in model in Eq.9 an external variable, $X_4 \sim N(0,1)$ with $\rho(X_1, X_4) = 0.99$, was considered. Given the instance $\boldsymbol{x}^* = \{0.1, 0.1, 0.1, 1\}$, the resulting ranking should not contemplate any contribute for $X_4$. As illustrated by Figure 9, both the δ and Shapley methods assign a null coefficient to $X_4$. Furthermore, δ-XAI results are coherent with the scenario without correlation (Figure 5). Shapley values are still able to detect $x_1$ as the most impacting feature analogously to the case without correlation (Panel B of Figures 10 and 5). However, when $\rho(X_1, X_4) = 0.99$ is introduced, Shapley values consider $x_3$ less important (and with a negative effect) than $x_2$ instead of giving them the similar weights as done by the δ-XAI.

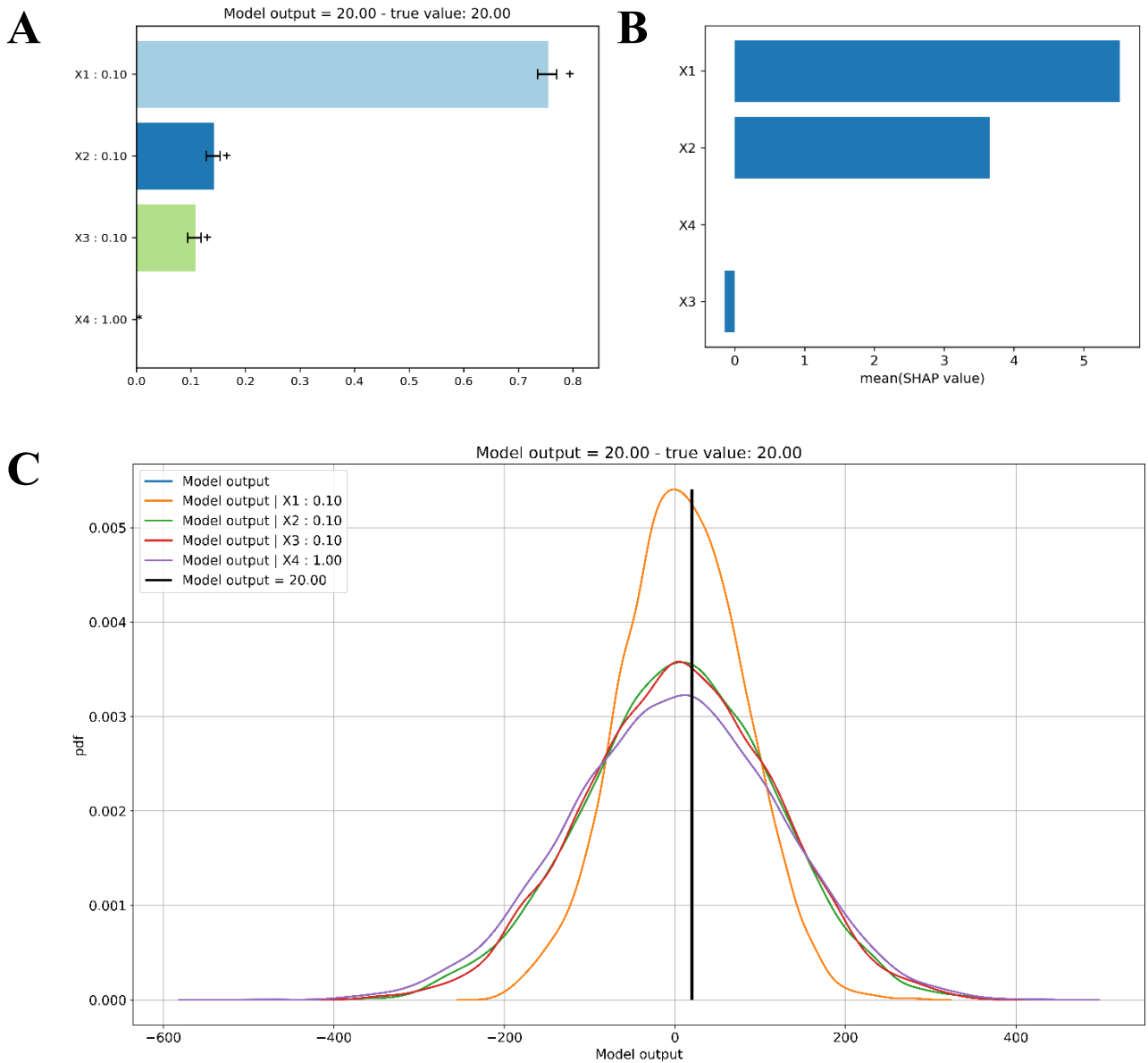

**Figure 9** Comparison between the δ indices and Shapley values on assessing features impact in predicting $\mathbf{x}^* = \{0.1,0.1,0.1,1\}$ with Eq.9 and given $\rho(X_1, X_4) = 0.99$. Panel A shows median and IQR (black bars) of the $\widehat{\delta_i}$ computed for each feature. A +/-/* symbol follows the bars depending by the signs of the $\delta_i$ in the computed with the bootstrap procedure. Panel B illustrates the ranking obtained with Shapley values. Panel C contains a graphical representation of how the $\delta_i$ computed for each feature value of $\mathbf{x}^*$.

Another experiment was conducted to assess how feature ranking is affected in the presence of two highly correlated model predictors. Considering the regression formula in Eq.9 and assuming that the most impacting variable, $X_1$, is highly correlated to $X_2$, with $\rho(X_1, X_2) = 0.99$. δ and Shapley values were computing for explaining the prediction of $\boldsymbol{x}^* = \{0.1, 0.1, 0.1\}$. As shown in Figure 10, the ranking of the features is the same as the condition without correlation (Figure 5). However, the δ-XAI value of $X_2$ is higher due to the strong correlation with $X_1$. Differently, Shapley increases the importance of $X_3$.

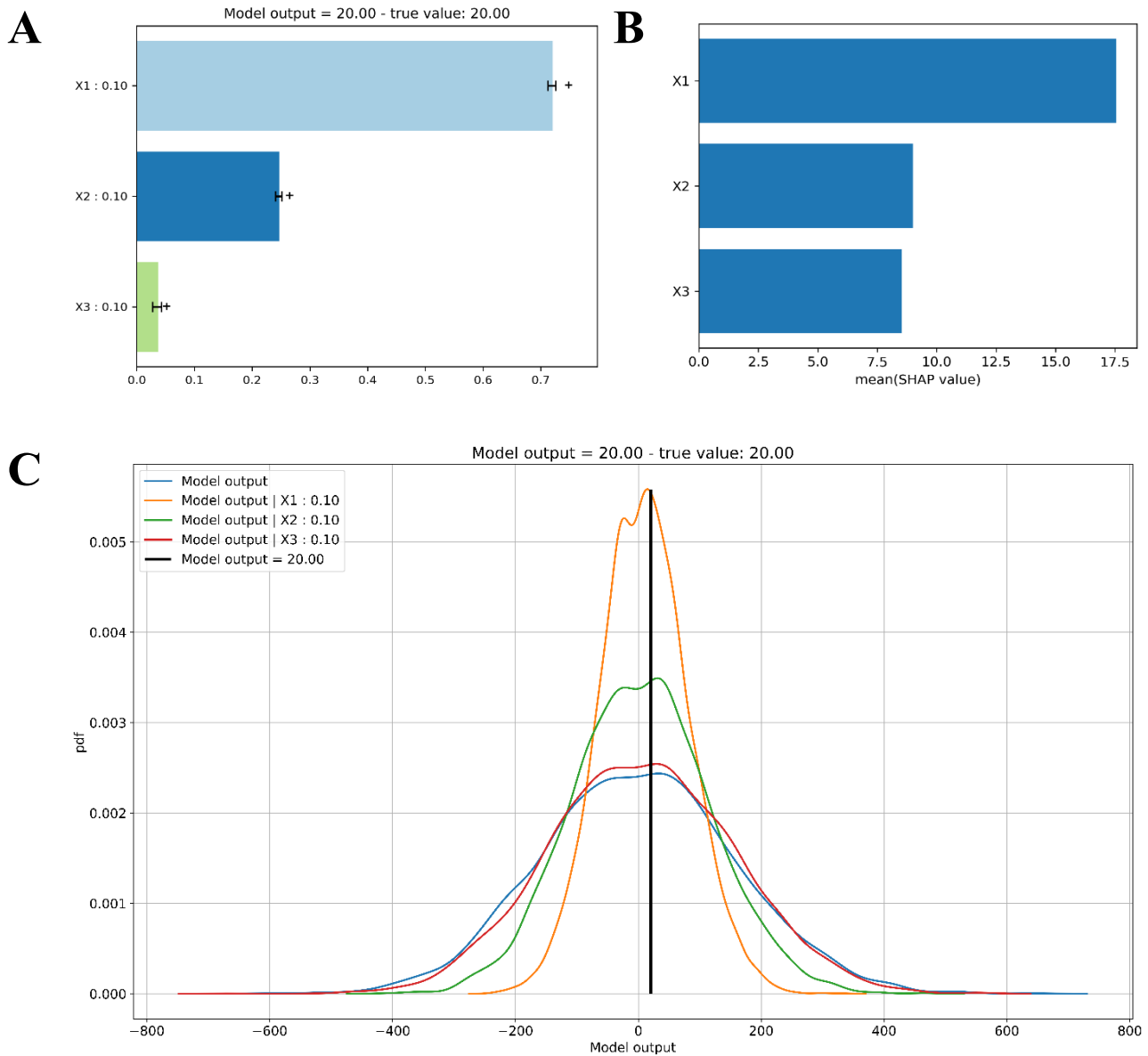

**Figure 18.0** Comparison between the δ indices and Shapley values on assessing features impact in predicting $\mathbf{x}^*$ = {0.1,0.1,0.1} with Eq.9 and given $\rho(X_1, X_2) = 0.99$. Panel A shows median and IQR (black bars) of the $\widehat{\delta_1}$ computed for each feature. A +/-/* symbol follows the bars depending by the signs of the $\delta_i$ in the computed with the bootstrap procedure. Panel B illustrates the ranking obtained with Shapley values. Panel C contains a graphical representation of how the $\delta_i$ computed for each feature value of $\mathbf{x}^*$.

# Discussions and Conclusion

XAI is currently at the core of the debate as it could facilitate the integration of AI/ML algorithms within the clinical practice. Indeed, the most outperforming AI/ML models (e.g., ensemble learners, deep neural networks) lack interpretability, thus preventing clinicians from fully trusting their predictions [2,3]. Currently, Shapley values are the most popular tool among AI practitioners in the healthcare domain. However, different works have highlighted that Shap values can lead to misleading insights on feature importance ranking

[11,33,34]. Therefore, some corrections to Shapley values were proposed and alternative XAI techniques are currently under development [7,8,12].

In particular, the adaptation of sensitivity analysis (SA) and global sensitivity analysis (GSA) to the XAI domain has recently gained momentum. Indeed, SA and GSA are intrinsically able to rank model inputs according to their impact on the prediction [17,32].

In this paper, we presented a novel approach called δ-XAI method to provide local explanations of ML model predictions leveraging the SA and GSA techniques. Indeed, the δ-XAI was defined by extending the δ index, a GSA metric exploited to assess the impact of each model parameter on the output probability density function [28]. The δ-XAI index was formalized to locally (e.g., for each specific instance) assess the impact of each feature's value on the predicted output (i.e., both for regression and classification problems) of a supervised ML model. In particular, the δ-XAI index quantifies how much a feature's value increase/decrease the probability of obtaining a given model prediction. Then, the pseudo-code describing the numerical implementation of the δ-XAI method was provided (Listing 1).

To better understand its performances, the proposed δ-XAI index was evaluated on simulated scenarios to locally explain the predictions of linear regression models as they are interpretable by design. Furthermore, Shapley values were used to benchmark each application of the δ-XAI index. The obtained results showed that:

- The δ-XAI index is overall coherent with Shapley values. The highest discrepancies between these methods were found in three scenarios: models with a highly impacting feature (Figure 8) and in the presence of extreme feature values (Figure 6). In presence of strong correlations, both methods returned similar features rankings correctly detecting both the most important feature and the irrelevant one (i.e., having a null coefficient in the regression model).

- In the presence of a model giving very high importance to a single feature (Eq.10), the δ-XAI index is more prone to rank it as the most impacting variable on model prediction, even if it apparently has no contribution (e.g., set to 0 in the linear model of Eq.10). This behavior is very interesting as we showed that it happens with Shapley values too but only when the gap between the main model feature and the others is very large (Figure 8). Therefore, the δ-XAI index showed a higher sensitivity in detecting dominant features in the model than Shapley values.

- Compared to the Shapley values, the δ-XAI index is more sensitive to extreme values of the features (i.e., far from the typical value in the population). This characteristic emerged in the scenarios illustrated in Figures 6 and 7, where it was shown that, in the absence of a strongly dominant feature in the model (Eq.9), the δ-XAI index can provide a features ranking that allows to identify those features that have a higher impact on model prediction due to their extreme values. This is an intrinsic characteristic of this methodology as it leverages feature distributions in their domains to compute a feature ranking. Therefore, the δ-XAI index is potentially able to detect distributional shifts or to check whether the ML model gives the adequate weight to rare conditions in its predictions. This latter aspect is of great value for a potential application of the δ-XAI within the clinical domain. Indeed, in such context some rare conditions (e.g., obesity, mutations) play a crucial role in the decision-making processes.

- From a qualitative perspective, it is our opinion that the δ-XAI provides more intuitive explanations than Shapley values. Indeed, for each instance, the effect of a feature's value on observing a certain model prediction is described by leveraging the basic concepts of probability density functions. In addition to the numerical values of the δ-XAI index, the obtained features ranking can be easily justified by using the plots shown in Panels C of Figures 4-10. Thus, differently by Shapley values, the δ-XAI method provides a clearer and "more explainable" features ranking to practitioners.

Overall, the obtained results highlighted that the δ-XAI is a promising method to robustly obtain local explanations of ML model predictions. Further investigations on real case studies will be performed to assess its impact on the AI-assisted clinical workflow.